\documentclass[twoside]{article}

\usepackage[preprint]{aistats2027}
\usepackage[round]{natbib}

\usepackage{amsmath,amssymb,mathtools}
\usepackage{booktabs}
\usepackage{array}
\usepackage{graphicx}
\usepackage{xcolor}
\usepackage[expansion=false]{microtype}
\usepackage{url}
\usepackage{placeins}
\usepackage{float}

\newcommand{\E}{\mathbb{E}}
\newcommand{\Pp}{\mathbb{P}}
\newtheorem{theorem}{Theorem}
\newtheorem{proposition}{Proposition}
\newtheorem{corollary}{Corollary}
\newtheorem{lemma}{Lemma}

\begin{document}

\runningtitle{Self-Consistency from Incomplete Logs}
\runningauthor{Gui, Chen, and Shen}

\twocolumn[
\aistatstitle{Which Answers Agree? \\ Evaluating Self-Consistency from Incomplete Logs}
\aistatsauthor{Zhengtao Gui \qquad Zexi Chen \qquad Weining Shen}
\aistatsaddress{University of California, Irvine}
]

\begin{abstract}
Self-consistency answers a question by generating multiple responses and
selecting the most frequent answer. Its evaluation requires both correctness
labels and information about which responses agree; correctness labels alone
generally determine neither voting accuracy nor the benefit of additional
votes. For tasks in which all correct responses map to the same normalized
answer, we derive sharp accuracy bounds using classical stochastic
majorization, refined by abstention counts and answer-category limits. We also
derive sharp bounds on accuracy changes across voting budgets that share a
response pool. When answer-category counts are available, we introduce
Plurality-U, a count-based dynamic program that exactly averages voting
accuracy over all fixed-size subsets, eliminating finite-pool Monte Carlo
error. On 1,403,520 stored outputs from recent reasoning models in Scorio Lite and
Scorio Trace, recording abstention counts and answer options cuts the width of
SuperGPQA accuracy intervals by more than half. For moving from 9 to 17 votes,
our bounds reduce interval width by a median of 77.7\% on SuperGPQA and 72.3\%
on 14 GPQA and Adaptive-Consistency pools, compared with a baseline that
combines separate accuracy bounds. Exact computation has lower paired median runtime
than 2,000-subset Monte Carlo in 14 of 15 tested pool--budget settings. These
results clarify what incomplete logs can establish and provide an exact
evaluation procedure when answer counts are retained.
\end{abstract}

\section{Introduction}

Self-consistency answers a question by generating several responses and
selecting the most frequent answer \citep{wang2023selfconsistency,saparkhan2026self}.
A natural question is how often this procedure returns the correct answer for a given voting budget. Existing work evaluates this accuracy from stored response pools by averaging over enumerated or sampled subsets
\citep{hassid2025dontoverthink,bahuguna2026backfires, hariri2026testtime,hariri2026ranking}. What can be inferred, however, depends on what the logs retain. Correctness labels support Pass@$k$
\citep{chen2021evaluating,brown2024monkeys} and the probabilities that a majority of sampled responses are correct, but they do not show whether wrong responses concentrate on the
same answer. As a result, identical correctness records can imply different voting accuracies.

For example, consider a pool of 20 valid responses, eight of which give the
same correct answer. Suppose we draw nine responses uniformly without replacement and
break voting ties uniformly. If all 12 wrong responses agree, expected voting
accuracy is 20.5\%; if they all give different answers, it is 97.8\%. The correctness
records are identical, but the voting outcomes are not.

Additional information about the response pool can narrow this uncertainty. In particular, answer-category limits and invalid-response counts rule out some wrong-answer allocations. We ask how precisely voting accuracy
can be identified as more information is retained. For tasks whose correct responses normalize to one answer, we derive sharp
finite-pool accuracy bounds, refined by abstention counts and an upper limit on
the number of answer categories. Every compatible accuracy lies in the stated
interval, and both endpoints are attainable. Classical stochastic majorization
characterizes the extremal allocations under category limits, while abstentions are
handled separately because they consume generation slots without casting a
vote.

Evaluation also raises a budget-selection question
\citep{byerly2026self,liu2026twocalls,wunderlich2026multi}: does using more
votes improve accuracy? Because both budgets draw from the same response pool, they depend on the same unknown allocation of wrong responses across answer categories. We therefore evaluate both budgets under every compatible allocation and directly optimize
their accuracy difference, yielding sharp bounds on the effect of additional
votes. This joint analysis is necessary because allocations that extremize one
budget's accuracy need not extremize the difference between budgets.

When complete answer-category counts are retained, the answer histogram and
voting rule uniquely determine finite-pool voting accuracy. We introduce
Plurality-U, a complete U-statistic, together with a count-based dynamic program for computing it exactly without enumerating all fixed-size subsets. The computation requires only the recorded histogram and eliminates Monte Carlo error from subset sampling. Under iid generation, Plurality-U is also unbiased
for the accuracy of voting on fresh responses.

We validate the framework on the recent Scorio Lite and Scorio Trace candidate
banks, comprising 1,403,520 attempts in 17,544 complete pools from 24 model configurations
across SuperGPQA and nine competition-mathematics benchmark splits
\citep{hariri2026testtime,hariri2026ranking,mapteam2025supergpqa}. Directly
bounding the accuracy change from 9 to 17 votes reduces interval width by a
median of 77.7\% on SuperGPQA and 72.3\% across 14 GPQA and Adaptive-Consistency pools, compared with combining separately optimized accuracy bounds. Exact computation has lower paired median runtime than 2,000-subset Monte Carlo in 14 of 15 tested pool--budget settings. We also evaluate sensitivity to answer normalization and invalid-response handling.

We summarize our contributions as follows:

\begin{itemize}
  \item \textbf{Evaluation under incomplete logs.} We derive sharp voting-accuracy bounds and characterize how answer-category limits and abstention counts refine them.
  \item \textbf{Comparisons across voting budgets.} We obtain sharp bounds on accuracy changes by evaluating different budgets over a shared underlying response pool.
  \item \textbf{Exact evaluation from answer counts.} We develop Plurality-U and its count-based dynamic program, and empirically evaluate information loss, computational cost, and sensitivity to evaluation rules.
\end{itemize}

\section{Related Work}

\paragraph{Metrics and selectors for repeated generation.}
Several metrics and answer-selection rules have been proposed for multiple responses to the
same question \citep{li2026confidence}. Pass@$k$ measures the probability that at least one of $k$ responses is correct
\citep{chen2021evaluating,brown2024monkeys}, while G-Pass@$k$ evaluates
stability through thresholds on the number of correct responses
\citep{liu2025stable}. Bayesian alternatives estimate average success or
rubric scores \citep{hariri2025dontpass}. Self-consistency instead selects
the most frequent answer \citep{wang2023selfconsistency}, with adaptive
variants stopping once agreement is sufficiently high
\citep{aggarwal2023adaptive}. Its accuracy therefore depends on both correctness and agreement among responses.

\paragraph{Finite-pool evaluation of voting policies.}
Stored response pools allow voting budgets to be compared without additional generation.
\citet{hassid2025dontoverthink} enumerate subsets of 20 generations and
average their voting outcomes. \citet{bahuguna2026backfires} study plurality
accuracy across budgets on GPQA, using enumeration or Monte Carlo sampling,
and investigate budget selection with complete answer logs.
\citet{shah2026upskill} distinguish plurality@$k$ from strict consensus
and evaluate voting on fixed response batches. We compute the same subset-based
expectation exactly from answer counts and characterize what can be inferred from incomplete logs
about voting accuracy and the benefit of additional votes.

\paragraph{Voting accuracy and answer agreement.}
Classical work bounds majority or multiclass plurality accuracy under dependent or independent-voter models
\citep{kuncheva2003limits,aeeneh2025new}. For language models, recent work examines correctness across voting budgets \citep{liu2026twocalls},
consensus among wrong answers \citep{zhang2026decomposing}, and sampling or selection
rules that distinguish total correct probability from its concentration on individual
answers \citep{yang2026correctmass}. A separate line of work asks whether sampled
votes recover the population mode, through sample allocation, certification,
or plug-in estimation
\citep{feng2025optimal,corderoencinar2025certified,nowak2025estimating}. Our setting differs in that we condition on a fixed stored response pool and ask how often a given finite-budget voting rule selects the correct answer.

\paragraph{Partial identification and exact computation.}
Partial identification characterizes the range of target values compatible with incomplete information, allowing decisions even when the target is not point identified
\citep{tamer2010partial,stoye2009partial}. We apply this framework to
correctness-only logs, treating different voting budgets as sharing the same unknown answer allocation. To incorporate category limits, we use the preservation of Schur order under hypergeometric sampling \citep{marshall2011inequalities}.

With complete answer counts, averaging over all fixed-size subsets gives a
complete U-statistic, whereas incomplete U-statistics evaluate only sampled
subsets \citep{hoeffding1948class,blom1976some,clemencon2016scaling}.
\citet{hazon2012evaluation} use dynamic programming to compute winner probabilities
under independent heterogeneous ballots; our dynamic program instead averages over uniform sampling without replacement from a fixed stored pool. Together, these two approaches cover different levels of retained information: partial identification gives sharp bounds when answer counts are missing, while complete counts allow exact evaluation. Table~\ref{tab:comparison} summarizes their inputs, targets and computation.

\begin{table*}[!t]
\caption{Related evaluation formulations, their required inputs, targets,
and computation. $a$ is the abstention count and $q$ the cap on wrong answer
categories (Section~4).}
\label{tab:comparison}
\centering
\small
\begin{tabular}{>{\raggedright\arraybackslash}p{0.24\textwidth}>{\raggedright\arraybackslash}p{0.18\textwidth}>{\raggedright\arraybackslash}p{0.26\textwidth}>{\raggedright\arraybackslash}p{0.22\textwidth}}
\toprule
Formulation & Retained input & Returned quantity & Computation \\
\midrule
\citet{chen2021evaluating,liu2025stable} & Correctness labels & Coverage or correctness-threshold score & Function of the correct count \\
\citet{hassid2025dontoverthink,bahuguna2026backfires} & Answer identities & Expected subset voting score & Enumeration or Monte Carlo \\
\citet{shah2026upskill} & One answer batch & Batch plurality score & Vote on that batch \\
\citet{hazon2012evaluation} & Uncertain-ballot model & Election-winner probabilities & Probabilistic voting DP \\
\citet{marshall2011inequalities} & Hypergeometric population counts & Schur ordering of expected payoffs & Preservation theorem \\
This work: incomplete logs & Labels; optionally $a,q$ & Sharp voting-accuracy bounds & Extremal allocations and DP \\
This work: full counts & Answer-count histogram & Exact all-subset voting accuracy & Count DP (complete U-statistic) \\
\bottomrule
\end{tabular}
\end{table*}

\section{Problem setup}

We evaluate a voting rule that takes $k$ model outputs and returns the
answer category with the most votes. We consider its accuracy in two
settings: on $k$ fresh outputs and on a uniformly sampled size-$k$ subset of a
fixed pool of $n$ stored outputs, where $1\leq k\leq n$. To formalize these
settings, fix a benchmark problem and a deterministic answer-normalization
rule. Let
$X_1,\ldots,X_n\in\{0,1,\ldots,m\}$ denote the normalized categories of the
stored outputs. Category 0 is the unique gold-answer category, while categories
$1,\ldots,m$ are incorrect. Normalization is label-blind: correctness labels
are used to score the selected answer, but not to decide whether two outputs
belong to the same category. We write
$C_j=\sum_i\mathbf{1}\{X_i=j\}$ and $C=(C_0,\ldots,C_m)$ for the resulting
answer-count histogram.

The selector returns a category with the largest count among the $k$
selected outputs and breaks ties uniformly. Its correctness payoff is
\begin{equation}
h_k(x_1,\ldots,x_k)=
\frac{\mathbf{1}\{0\text{ has maximal count}\}}
{\#\{\text{categories with maximal count}\}}.
\label{eq:kernel}
\end{equation}
Thus, $h_k=1$ when the gold category is the unique winner, $h_k=0$ when it is
not a winner, and $h_k=1/t$ when it is tied with $t-1$ other categories.

The two evaluation settings define two related targets. If fresh outputs are
drawn iid from a fixed distribution $p$, their voting accuracy is
$\theta_k(p)=\E_p[h_k(X_1,\ldots,X_k)]$. For the fixed stored pool, we instead
average the payoff over every size-$k$ subset:
\begin{equation}
U_{n,k}(C)=\binom nk^{-1}\sum_{S\subset[n]:|S|=k}h_k(X_S).
\label{eq:u}
\end{equation}
The finite-pool quantity $U_{n,k}(C)$ is the exact expected payoff of a
uniformly drawn $k$-subset and
requires no probabilistic assumption on the stored pool. If the stored outputs
are iid draws from $p$, then $U_{n,k}(C)$ is also an unbiased estimator of the
fresh-sample target $\theta_k(p)$. Our finite-pool calculations and
identification bounds concern $U_{n,k}(C)$ and therefore do not require the iid
assumption.

Correctness-only metrics discard information needed to determine plurality
accuracy. If $R$ is the number of correct outputs in the selected set,
strict-majority correctness is $\mathbf{1}\{R>k/2\}$ and Pass@$k$ is
$\mathbf{1}\{R\geq1\}$. Both are functions of $R$ alone, whereas plurality
accuracy also depends on how the remaining $k-R$ outputs are allocated among
wrong answer categories.

\section{Identification from incomplete logs}

We begin with the coarsest log, which records only the pool size $n$ and the number
$c=C_0$ of correct outputs, but not how the $n-c$ incorrect outputs are grouped
into answer categories. Assume for now that every output casts a valid vote.
For a uniformly selected size-$k$ subset, the number of correct outputs obeys
$R\sim\operatorname{Hypergeom}(n,c,k)$. Define
\begin{align}
L_{n,c,k}&=\Pp(R>k-R)+\tfrac12\Pp(R=k-R),\label{eq:lower}\\
B_{n,c,k}&=\Pp(R\geq2)+\tfrac1k\Pp(R=1).\label{eq:upper}
\end{align}

\begin{theorem}[Sharp correctness-only bounds]
Over all allocations of the $n-c$ incorrect outputs to answer categories, the
finite-pool expected plurality accuracy lies in
$[L_{n,c,k},B_{n,c,k}]$, and both endpoints are attainable.
\label{thm:bounds}
\end{theorem}

The endpoints correspond to two extreme error allocations. Concentrating all
incorrect outputs in one category creates the strongest competitor to the gold
answer and attains $L_{n,c,k}$. At the other extreme, assigning every incorrect output to its own
category disperses the wrong votes and attains $B_{n,c,k}$. Splitting a wrong
category can only weaken its competition with the gold category, so no allocation can yield accuracy outside these bounds. The full proof is in Appendix~\ref{app:proofs}. As a special case, for
an odd $k$ and a binary answer alphabet, all incorrect outputs necessarily share one category. The bounds therefore coincide, and plurality accuracy reduces to strict-majority correctness.

\subsection{Known answer-alphabet size}

The preceding upper bound allows every incorrect output to form its own
category. If the task is independently known to admit at most $q\geq1$ wrong
answer categories \citep{xia2024sportqa,guo2025fineval}, this allocation may be infeasible and the upper bound can
tighten. Let $V_k(c,w)$ denote finite-pool plurality accuracy for $c$ correct
outputs and wrong-count vector $w$. For $e=n-c$, let $b_q(e)$ be the length-$q$
vector whose entries sum to $e$ and differ by at most one; zero entries
represent unused categories.

\begin{theorem}[Known alphabet]
For $1\leq k\leq n$, over all wrong-count vectors with at most $q$ categories,
the minimum is $L_{n,c,k}$ and the maximum is
\[
B^{(q)}_{n,c,k}=V_k\bigl(c,b_q(n-c)\bigr).
\]
Both extrema are attainable.
\label{thm:boundedq}
\end{theorem}

The maximizing allocation $b_q(e)$ spreads the incorrect outputs as evenly as
the category limit permits. This follows from classical stochastic
majorization under hypergeometric sampling
\citep{nevius1977schur,marshall2011inequalities}. Equivalently, pairwise
smoothing moves one physical wrong output from a category of size $a$ to one
of size $b$ when $a\geq b+2$ and cannot decrease the expected gold payoff.
Repeating the operation produces $b_q(e)$, and
Appendix~\ref{app:smoothing} treats all tie cases. When $q=1$, the two
endpoints coincide. For $q\geq n-c$, the category limit no longer binds, and
the upper endpoint reduces to Eq.~\eqref{eq:upper}.

\subsection{Abstention-aware bounds}

Some logs distinguish invalid outputs that consume a generation slot but do
not cast a vote. Let $a$ be the number of these abstentions and $N=n-a$ the
number of valid outputs. We assign zero payoff when all selected outputs
abstain. Conditional on selecting $v\geq1$ valid outputs, they form a uniform
size-$v$ subset of the valid pool. The probability of this event is
\[
w_v=\frac{\binom Nv\binom a{k-v}}{\binom nk}
\]
for $\max(1,k-a)\leq v\leq\min(k,N)$.

\begin{corollary}[Observed abstentions]
Given $(n,c,a,q,k)$, the sharp lower and upper endpoints are
\begin{align*}
L^a_{n,c,k}&=\sum_v w_v L_{N,c,v},\\
B^{a,q}_{n,c,k}&=\sum_v w_v B^{(q)}_{N,c,v}.
\end{align*}
With no cap $q$, replace $B^{(q)}$ by $B$. If $a$ is also unrecorded and
abstentions are allowed, the sharp endpoints become
$[L_{n,c,k},\Pp(R\geq1)]$.
\label{cor:abstention}
\end{corollary}

The bounds are weighted averages of the corresponding conditional bounds for
each possible number $v$ of valid votes. The same concentrated and balanced
allocations attain the conditional endpoints for every $v$, so the resulting
bounds remain sharp. If $a$ is unrecorded, treating every noncorrect output as
an abstention attains the upper endpoint. Accounting for abstentions can substantially change the
identified interval. For example, with one gold output, two abstentions, and
$k=3$, the gold answer wins with probability one, whereas applying the
no-abstention bound would give an upper bound of only  $1/3$. Empty sums when $N=0$ equal zero.

\section{Exact evaluation across budgets}

With the complete answer-count histogram $C$, the finite-pool accuracy in
Eq.~\eqref{eq:u} is point identified and exactly computable. We call the
resulting complete U-statistic Plurality-U.

\begin{theorem}[Unbiased estimation]
For $n\geq k$ and iid stored outputs from $p$,
$\E[U_{n,k}]=\theta_k(p)$. Let $S$ be an independent,
uniform $k$-subset and put $T=h_k(X_S)$.  Then
\begin{align*}
U_{n,k}&=\E[T\mid X_1,\ldots,X_n],\\
\operatorname{Var}(U_{n,k})&\leq\operatorname{Var}(T).
\end{align*}
\label{thm:ustat}
\end{theorem}

Every fixed $k$-subset has distribution $p^k$, which gives unbiasedness. The
conditional-expectation identity and the law of total variance give the
variance bound.
Appendix~\ref{app:proofs} gives the UMVU refinement for a fixed finite answer
alphabet. Abstentions use the corresponding kernel, equivalently
$\sum_v w_v U_{N,v}(C_{\mathrm{valid}})$. More generally, unbiasedness and
variance reduction require only that the evaluation rule be fixed, bounded,
and invariant to the order of its $k$ inputs, so they are not specific to
plurality voting. By contrast, the identification bounds in Section~4 require
all correct outputs to belong to one answer category.

The complete statistic also permits several normalized correct categories.
Suppose categories are constructed without correctness labels. Category $j$
contains $C_j$ outputs, of which $g_j$ are correct. After selecting a winning
category, the rule returns one of its selected members uniformly. Category
$j$ wins with probability $V_k(C_j,C_{-j})$, and its returned member is correct
with expected probability $g_j/C_j$. Hence the exact expected score is
\[
\sum_j \frac{g_j}{C_j}V_k(C_j,C_{-j}),
\]
over nonempty categories. Correctness labels affect only scoring, not category
construction or winner selection, so the complete-U argument remains valid.

Exact evaluation need not enumerate the $\binom nk$ physical subsets in
Eq.~\eqref{eq:u}. We use a count-based dynamic program. Fix $r$, the selected
gold-category count. After processing $j$ wrong categories, $D_j(s,t)$ counts
selections with $s$ wrong outputs, no wrong count above $r$, and exactly $t$
wrong categories tied with the gold category at $r$.
Starting at $D_0(0,0)=1$, a category with pool count $C_j$ gives
\begin{equation}
D_j(s+\ell,t+\mathbf{1}\{\ell=r\})\mathrel{+}=
D_{j-1}(s,t)\binom{C_j}{\ell},
\label{eq:dp}
\end{equation}
for $0\leq\ell\leq\min(C_j,r,k-r-s)$. After all $m$ wrong categories, the
exact numerator is
\begin{equation}
\sum_{r=1}^{\min(C_0,k)}\binom{C_0}{r}
\sum_t\frac{D_m(k-r,t)}{t+1}.
\label{eq:numerator}
\end{equation}
Dividing by $\binom nk$ yields Eq.~\eqref{eq:u}. States track category counts,
not output identities, so the implemented recurrence is polynomial in the
number of observed categories and $k$. Concentrated histograms yield few
nonzero states and are especially cheap. Appendix~\ref{app:algorithm} proves
the recurrence and gives its complexity.

\subsection{Shared-allocation budget contrasts}
\label{sec:budgetbounds}

To determine whether a larger budget helps, both accuracies must use the same
unknown allocation of incorrect outputs. Fix $(n,c,a,q)$, let $e=n-c-a$, and
write the compatible allocations as
\[
\mathcal W_{e,q}=\{w\in\mathbb Z_{\geq0}^q:
 w_1\geq\cdots\geq w_q,\ \textstyle\sum_jw_j=e\}.
\]
Each $w\in\mathcal W_{e,q}$ is one allocation consistent with the log. Let
$F_k(w)=\sum_v w_v^{(k)}V_v(c,w)$ be its abstention-aware payoff, where
$w_v^{(k)}$ is the probability of obtaining $v$ valid votes at budget $k$.
Using the same all-abstention convention, define the budget gain as
$\Delta_{k,\ell}(w)=F_\ell(w)-F_k(w)$ for $k<\ell\leq n$.

\begin{proposition}[Budget contrasts]
\label{prop:contrast}
Given $(n,c,a,q)$, optimizing over the same allocation at both budgets gives
the sharp gain endpoints
\begin{align*}
 d^-_{k,\ell}&=\min_{w\in\mathcal W_{e,q}}\Delta_{k,\ell}(w),\\
 d^+_{k,\ell}&=\max_{w\in\mathcal W_{e,q}}\Delta_{k,\ell}(w).
\end{align*}
If $[L_j,B_j]$ are the sharp single-budget bounds, then
\[
[d^-_{k,\ell},d^+_{k,\ell}]
 \subseteq [L_\ell-B_k,B_\ell-L_k].
\]
For a fixed benchmark with no cross-item constraints on missing allocations,
macro gain endpoints are the averages of the item endpoints.
\end{proposition}

We enumerate $w\in\mathcal W_{e,q}$, evaluate both budgets by count DP, and
retain the smallest and largest differences. Using the same $w$ excludes
marginal extrema that require incompatible allocations. Because the set is
finite, the optimizers also show that the endpoints are sharp. There are at most
$\binom{e+q-1}{q-1}$ allocations, and items with the same $(n,c,a,q)$ share
them, so enumeration remains practical for ten answer options (Section~6.3).
Appendices~\ref{app:contrastproof} and~\ref{app:allbudget} give the proof and an all-budget polynomial recurrence.

Gain extrema need not occur at the concentrated or balanced allocations. For
$n=7,c=2,a=0,q=3$, comparing $k=3$ with $\ell=7$, allocations $(5,0,0)$ and
$(2,2,1)$ give gains $-1/7$ and $4/105$, but $(3,1,1)$ gives the smaller value
$-29/105$. Differences of Schur-concave payoffs need not preserve their
ordering. Thus $d^-_{k,\ell}>0$ certifies improvement, $d^+_{k,\ell}<0$
certifies decline, and otherwise the direction is unidentified.

\section{Experiments}
\label{sec:experiments}

We first describe the setup (\S\ref{sec:policies}). We then report three
experiments: one measuring how much each extra field in the log narrows the
accuracy interval (\S\ref{sec:exp-info}), one testing whether a shared
allocation can tell which voting budget is better (\S\ref{sec:exp-budget}),
and one measuring how sensitive the results are to answer processing
(\S\ref{sec:exp-policy}). Finally, \S\ref{sec:exp-compute} compares exact evaluation with Monte
Carlo sampling.

\subsection{Setup}
\label{sec:policies}

\paragraph{Data.} Table~\ref{tab:data} lists the stored outputs used in each
experiment. Our main data are Scorio Lite and Scorio Trace
\citep{hariri2026testtime,hariri2026ranking}: 1,403,520 outputs from 24 recent
reasoning-model configurations on SuperGPQA \citep{mapteam2025supergpqa} and
nine competition-mathematics sets. We also use the GPQA and
Adaptive-Consistency pools
\citep{rein2023gpqa,bahuguna2026backfires,aggarwal2023adaptive}. Nine of the
ten Adaptive-Consistency tasks have fixed options. Among their 18 pools, 12 have
three to six options and 6 are binary. The MATH pools
\citep{hendrycks2021math} are used for the answer-processing and timing
studies.

\begin{table}[!tbp]
\caption{Stored outputs used in the experiments; $n$ is the number of stored
outputs per question.}
\label{tab:data}
\centering\scriptsize
\setlength{\tabcolsep}{2pt}
\begin{tabular}{@{}>{\raggedright\arraybackslash}p{1.55cm}>{\raggedright\arraybackslash}p{2.55cm}>{\raggedright\arraybackslash}p{2.05cm}r>{\raggedright\arraybackslash}p{0.85cm}@{}}
\toprule
Source & Models & Questions & $n$ & Used in \\
\midrule
Scorio Lite & Qwen3.6-35B-A3B; gpt-oss-20b (3 settings) & 3,600 SuperGPQA; 186 math & 80 & \S6.2--6.4 \\
Scorio Trace & 20 reasoning models & 120 math & 80 & \S6.2 \\
GPQA & Qwen2.5-7B, Llama-3-8B & 198 GPQA Diamond & 64 & \S6.2--6.5 \\
Adaptive-Consistency & code-davinci-002, Vicuna-13B & 10 tasks, 146--2,279 each & 40 & \S6.2--6.4 \\
MATH & Qwen2.5-Math-1.5B, its Oat-Zero variant & 500 MATH-500; 716 other MATH & 64 & \S6.4, 6.5 \\
\bottomrule
\end{tabular}
\end{table}

\paragraph{Answer processing.} We group outputs by the answers extracted in
each release, without using correctness labels. We apply Unicode normalization
and trimming, ignore letter case for multiple-choice answers, and treat
missing or out-of-range answers as abstentions. Ties are broken at random,
and a subset in which every output abstains scores zero. Unless stated otherwise, we use $k=9$
votes. Appendices~\ref{app:data}
and~\ref{app:independent} give the sources, rules and full results.

\begin{figure*}[!t]
\centering
\includegraphics[width=0.98\textwidth]{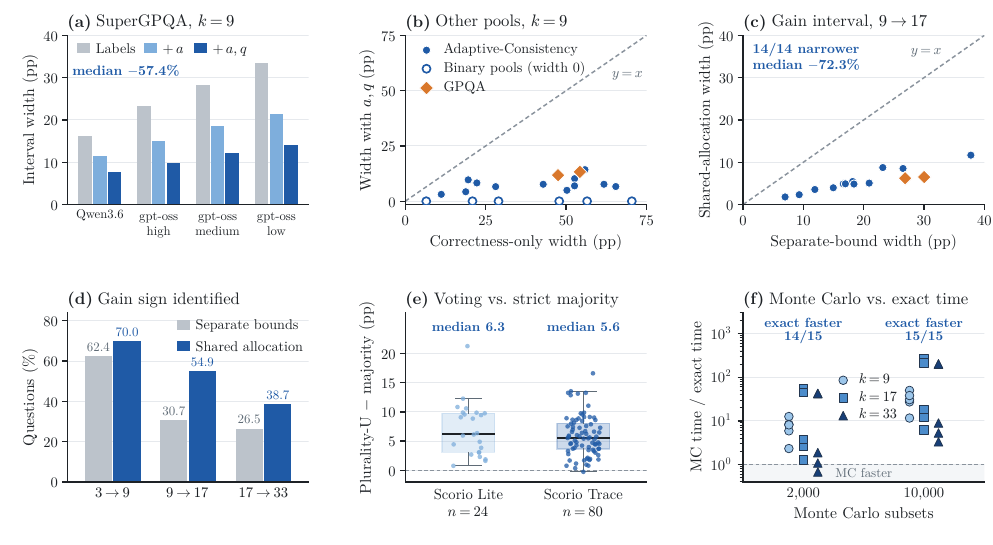}
\caption{\textbf{Main results.} (a) SuperGPQA interval width at
$k=9$ under three logs. (b) Width with correctness only versus with abstentions
and options, 20 GPQA and Adaptive-Consistency pools. (c) $9\!\to\!17$ gain-interval width, shared
versus separate bounds, 14 pools. (d) Share of 3,126 questions with a known
gain sign. (e) Plurality-U minus strict-majority correctness, 104 Scorio
cells. (f) Median Monte Carlo/exact time ratio, 30 settings.}
\label{fig:experimentsummary}
\end{figure*}

\begin{table}[!tbp]
\caption{SuperGPQA at $k=9$. Exact is voting accuracy computed from full
answer counts. The other columns give the identification interval width when
the log keeps correctness labels only, adds abstention counts ($+a$), or also
adds the option count, which caps the wrong categories ($+a,q$). Parentheses
give the reasoning effort.}
\label{tab:abstentionbounds}
\centering\footnotesize
\setlength{\tabcolsep}{3.1pt}
\begin{tabular}{@{}lrrrr@{}}
\toprule
 & & \multicolumn{3}{c}{Interval width (pp)} \\
\cmidrule(l){3-5}
Model & Exact (\%) & Labels & $+a$ & $+a,q$ \\
\midrule
Qwen3.6-35B-A3B & 65.95 & 16.28 & 11.44 & \textbf{7.80} \\
gpt-oss-20b (high) & 47.66 & 23.28 & 14.94 & \textbf{9.86} \\
gpt-oss-20b (medium) & 45.49 & 28.39 & 18.47 & \textbf{12.15} \\
gpt-oss-20b (low) & 39.07 & 33.48 & 21.47 & \textbf{14.04} \\
\bottomrule
\end{tabular}
\end{table}

\subsection{Information Retained in the Log}
\label{sec:exp-info}

Table~\ref{tab:abstentionbounds} shows how the retained information affects the width of the identified accuracy interval. With correctness labels alone, the log does not record whether incorrect outputs agree with one another. If they all share one answer, they form a strong competitor to the correct answer; if they are spread across different answers, the correct answer can win with fewer votes. On SuperGPQA, the resulting interval width ranges from 16.28 to 33.48 points, comparable to the accuracy differences between the four models.

Recording the number of invalid outputs yields the largest reduction. Without
this count, every incorrect output could in principle be invalid. Because invalid outputs do not vote, the correct
answer would then win whenever at least one correct output is selected, which raises
the upper endpoint (Corollary~\ref{cor:abstention}). Recording the count
excludes this case and reduces the width to 11.44--21.47 points. Recording
the number of answer options reduces it further to 7.80--14.04 points,
because a question with ten options admits at most nine distinct incorrect
answers. In total, the two counts reduce the width by a median of 57.4\%
(Figure~\ref{fig:experimentsummary}a).

The GPQA and Adaptive-Consistency pools show the same pattern
(Figure~\ref{fig:experimentsummary}b). For the two GPQA pools, the width
decreases from 47.50 and 54.27 points to 11.77 and 13.29 points, and for the
12 Adaptive-Consistency pools with three to six options it decreases from
11.20--65.44 to 3.11--14.28 points. The six binary pools become point identified: with
only two possible answers, all incorrect outputs necessarily share one answer,
so no uncertainty remains once invalid outputs are counted
(Theorem~\ref{thm:boundedq} with $q=1$). In all cases, the width reflects
missing information rather than sampling error. When the log records the
count of every answer, the interval reduces to a single value, which Plurality-U computes exactly.

The exact values also differ systematically from scores based on correctness
alone. Strict-majority correctness counts a question as solved when more than
half of the $k$ votes are correct. Under the single-gold-category setup of
Section~3, voting accuracy is never below this score,
because a correct answer that receives more than half of the votes always
wins. Voting accuracy is often higher, because the correct answer can also
win with fewer votes when the incorrect answers are split. Across the 104 Scorio
model--task cells, voting accuracy exceeds strict-majority correctness by a
median of 5.78 points (interquartile range 3.49--8.72;
Figure~\ref{fig:experimentsummary}e).

\subsection{Comparing Voting Budgets}
\label{sec:exp-budget}

To compare 9 and 17 votes, we consider the gain, defined as the accuracy with 17 votes minus that with 9 votes. With incomplete logs, the gain is identified only up to an interval, whose sign is known when the entire interval lies above or below zero. A simple baseline subtracts the single-budget intervals from Section 4, pairing the upper endpoint at 17 votes with the lower endpoint at 9 votes to obtain an upper bound on the gain. However, these endpoints generally correspond to different allocations of incorrect outputs (Theorems~\ref{thm:bounds} and~\ref{thm:boundedq}). Because both budgets are evaluated on the same stored pool, they must share a single allocation. Our shared-allocation bounds therefore optimize the gain directly over all allocations consistent with the log (Proposition~\ref{prop:contrast}).

Both methods use the same retained information: the numbers of stored, correct and
invalid outputs, and the number of answer options. We apply them to four
SuperGPQA pools, one for each Scorio Lite model with all 3,600 questions, and
to the 3,126 questions in the 14 GPQA and Adaptive-Consistency pools with
three or more options. We exclude the six binary Adaptive-Consistency pools because their incorrect outputs necessarily share one answer, making the gain point identified. We fixed $9\!\to\!17$ as the
main comparison, and $3\!\to\!9$ and $17\!\to\!33$ as secondary comparisons, before computing any gain.

Enforcing a shared allocation substantially narrows the gain intervals. For
$9\!\to\!17$ on SuperGPQA, the interval for the pool average is 77.5--78.0\%
narrower (median 77.7\%), and the sign of the gain, that is, whether more
votes help or hurt, becomes known for 3,613 of the 14,400 model--question
pairs (25.1\%), compared with 269 (1.9\%) under the baseline
(Table~\ref{tab:budgetsuper} in Appendix~\ref{app:budget}). On the other 14
pools, the interval for the pool average is 62.2--78.4\% narrower (median
72.3\%; Figure~\ref{fig:experimentsummary}c), and the sign becomes known for
1,717 of the 3,126 questions (54.9\%), compared with 959 (30.7\%) under the
baseline (Figure~\ref{fig:experimentsummary}d). The secondary comparisons
improve in the same way on both sets (Table~\ref{tab:budgetsummary}).

A benchmark, however, applies one budget to all of its questions, so the
decision depends on the average gain over a pool. The interval for this
average runs from the average of the question-level lower endpoints to the
average of the upper endpoints (Proposition~\ref{prop:contrast}). For all
four SuperGPQA pools and 13 of the other 14 pools, this interval for
$9\!\to\!17$ still contains zero, so the log cannot tell whether 17 votes beat
9 votes on average. The only exception is Vicuna-13B on Disambiguation QA,
whose lower endpoint is positive (0.237 points). The average gains themselves are small: computed from
the full logs, they range from 0.19 to 0.56 points on SuperGPQA and from
$-0.22$ to 3.33 points on the other pools (Tables~\ref{tab:allbudgetpools}
and~\ref{tab:budgetsuper}), and an interval must be very narrow to exclude a
value this close to zero. Thus, shared allocation identifies the direction of the gain for many more individual questions, but benchmark-level budget selection usually still requires full answer counts.

\subsection{Sensitivity to Answer Processing}
\label{sec:exp-policy}

Before voting, each output is converted into an answer, and invalid outputs
are handled by a fixed rule. These choices determine which outputs count as
the same answer. Exact evaluation removes the uncertainty from missing counts
but not the effect of these choices (Table~\ref{tab:policysensitivity} in
Appendix~\ref{app:normalizers}). On the MATH-500 outputs of the base model
Qwen2.5-Math-1.5B (base MATH), keeping only the last boxed expression instead
of parsing each output with a mathematical checker leaves 36.50\% of the
outputs without a usable answer and lowers accuracy from 54.92\% to 37.36\%.
Treating invalid outputs as one shared incorrect answer instead of as
abstentions changes accuracy by at most 0.30 points on SuperGPQA and GPQA,
where invalid outputs are rare, but by up to 49.11 points on
Adaptive-Consistency, where 22.64\% of the outputs are invalid and grouping
them creates a large block of identical incorrect votes. An exact value is
exact only for a given processing rule, so the rule should be reported with
it.

\subsection{Exact Evaluation versus Monte Carlo}
\label{sec:exp-compute}

Plurality-U is defined as the average vote over all size-$k$ subsets of the
stored pool (Eq.~\eqref{eq:u}). Prior work computes this average by
enumerating or sampling subsets
\citep{hassid2025dontoverthink,bahuguna2026backfires}. Enumeration is exact but
feasible only for small pools, whereas Monte Carlo sampling estimates the
average from randomly drawn subsets, so its estimate contains random error.
Plurality-U computes the average exactly without enumeration, and we compare
its error and runtime with Monte Carlo.

The timing study uses the five pools with 64 stored outputs per question:
the three MATH pools and the two GPQA pools (Table~\ref{tab:data}). From each
pool we take ten questions selected by a hash of their identifiers. For each
question and each budget $k\in\{9,17,33\}$, we time exact evaluation and
Monte Carlo 20 times each and compare their median times
(Appendix~\ref{app:operational}). With 2,000 random subsets, the root mean squared error of
Monte Carlo ranges from 0.00192 to 0.00639 across pools, and Monte Carlo is
slower than exact evaluation in 14 of the 15 pool--budget settings. With 10,000 subsets,
the error decreases to 0.00063--0.00282, and exact evaluation is faster in all
15 settings (Figure~\ref{fig:experimentsummary}f). The only exception at
2,000 subsets is base MATH at $k=33$, whose answers are spread over many
categories, which increases the number of states in the dynamic program
(Table~\ref{tab:runtimecells} in Appendix~\ref{app:operational}).

\section{Discussion and conclusion}

This work determines how precisely stored logs identify the accuracy of self-consistency. With correctness labels alone, voting accuracy and the
effect of additional votes are known only within wide intervals. Recording the
number of invalid outputs and answer options more than halves these intervals, while jointly accounting for the shared answer allocation across voting budgets further tightens bounds on the effect of additional votes. With complete answer counts, voting accuracy is point identified and can be computed exactly, usually at lower cost than Monte Carlo. Benchmarks should therefore release answer counts when possible, or report identification intervals when they are unavailable.

Two limitations suggest directions for future work.
First, the exact values describe the stored outputs. Interpreting them as estimates of voting accuracy on fresh responses requires iid generation under a fixed prompt and decoder. Second, our experiments cover knowledge questions, competition mathematics, and reasoning tasks, but not structured-output tasks such as proof autoformalization \citep{gui2026evaluating,jana2026proofbridge} or text-to-SQL generation \citep{yu2018spider,li2023bird}.

\FloatBarrier
\subsection*{AI use statement}

In this work, we used ChatGPT and Codex to polish the language of this paper, organize and improve parts of our code, organize data and relevant literature, and check whether parts of our proofs needed further clarification. We independently reviewed and verified all AI-assisted work, including the proofs, code, data, references, and manuscript revisions. The research ideas, mathematical claims, experimental design, and interpretation of results are the authors’ own. We take full responsibility for the final paper.

\bibliography{references}

\begin{thebibliography}{}

\bibitem[Aeeneh et~al., 2025]{aeeneh2025new}
Aeeneh, S., Zlatanov, N., and Yu, J. (2025).
\newblock New bounds on the accuracy of majority voting for multiclass
  classification.
\newblock {\em IEEE Transactions on Neural Networks and Learning Systems},
  36(4):6014--6028.

\bibitem[Aggarwal et~al., 2023]{aggarwal2023adaptive}
Aggarwal, P., Madaan, A., Yang, Y., and Mausam (2023).
\newblock Let's sample step by step: Adaptive-consistency for efficient
  reasoning and coding with {LLM}s.
\newblock In {\em Proceedings of the 2023 Conference on Empirical Methods in
  Natural Language Processing}.

\bibitem[Bahuguna, 2026]{bahuguna2026backfires}
Bahuguna, U. (2026).
\newblock When self-consistency backfires: Majority vote hurts the majority of
  hard science problems for small {LLM}s.
\newblock arXiv preprint arXiv:2608.11403.

\bibitem[Blom, 1976]{blom1976some}
Blom, G. (1976).
\newblock Some properties of incomplete {U}-statistics.
\newblock {\em Biometrika}, 63(3):573--580.

\bibitem[Brown et~al., 2024]{brown2024monkeys}
Brown, B., Juravsky, J., Ehrlich, R., Clark, R., Le, Q.~V., R\'{e}, C., and
  Mirhoseini, A. (2024).
\newblock Large language monkeys: Scaling inference compute with repeated
  sampling.
\newblock arXiv preprint arXiv:2407.21787.

\bibitem[Byerly and Khashabi, 2026]{byerly2026self}
Byerly, A. and Khashabi, D. (2026).
\newblock Self-consistency falls short! {The} adverse effects of positional
  bias on long-context problems.
\newblock {\em Transactions of the Association for Computational Linguistics},
  14:292--317.

\bibitem[Chen et~al., 2021]{chen2021evaluating}
Chen, M. et~al. (2021).
\newblock Evaluating large language models trained on code.
\newblock arXiv preprint arXiv:2107.03374.

\bibitem[Cl\'{e}men\c{c}on et~al., 2016]{clemencon2016scaling}
Cl\'{e}men\c{c}on, S., Colin, I., and Bellet, A. (2016).
\newblock Scaling-up empirical risk minimization: Optimization of incomplete
  {U}-statistics.
\newblock {\em Journal of Machine Learning Research}, 17(76):1--36.

\bibitem[Cordero-Encinar and Duncan, 2025]{corderoencinar2025certified}
Cordero-Encinar, P. and Duncan, A.~B. (2025).
\newblock Certified self-consistency: Statistical guarantees and test-time
  training for reliable reasoning in {LLM}s.
\newblock arXiv preprint arXiv:2510.17472.

\bibitem[Feng et~al., 2025]{feng2025optimal}
Feng, A., Alonso, M., and Odonnat, A. (2025).
\newblock Optimal self-consistency for efficient reasoning with large language
  models.
\newblock arXiv preprint arXiv:2511.12309.

\bibitem[Geva et~al., 2021]{geva2021did}
Geva, M., Khashabi, D., Segal, E., Khot, T., Roth, D., and Berant, J. (2021).
\newblock Did {A}ristotle use a laptop? {A} question answering benchmark with
  implicit reasoning strategies.
\newblock {\em Transactions of the Association for Computational Linguistics},
  9:346--361.

\bibitem[Gui et~al., 2026]{gui2026evaluating}
Gui, Z., Yang, S., and Shi, Z. (2026).
\newblock Evaluating the robustness of proof autoformalization in {Lean} 4.
\newblock arXiv preprint arXiv:2606.14867.

\bibitem[Guo et~al., 2025]{guo2025fineval}
Guo, X., Xia, H., Liu, Z., Cao, H., Yang, Z., Liu, Z., Wang, S., Niu, J., Wang,
  C., Wang, Y., Liang, X., Huang, X., Zhu, B., Wei, Z., Chen, Y., Shen, W., and
  Zhang, L. (2025).
\newblock {FinEval}: A {Chinese} financial domain knowledge evaluation
  benchmark for large language models.
\newblock In {\em Proceedings of the 2025 Conference of the Nations of the
  Americas Chapter of the Association for Computational Linguistics: Human
  Language Technologies (Volume 1: Long Papers)}, pages 6258--6292. Association
  for Computational Linguistics.

\bibitem[Hariri et~al., 2026a]{hariri2026testtime}
Hariri, M., Chen, W., Shahini, N., Singh, V., Ye, K., Samandar, A., Ganguly,
  D., Sankar, S., Zhang, Y., Wang, S., Peng, J., Zhang, B., Hinczewski, M., and
  Chaudhary, V. (2026a).
\newblock Test-time scaling in reasoning {LLM}s: Inference regimes, evaluation,
  and reproducibility.
\newblock arXiv preprint arXiv:2608.04001.

\bibitem[Hariri et~al., 2026b]{hariri2026ranking}
Hariri, M., Hinczewski, M., Ma, J., and Chaudhary, V. (2026b).
\newblock Ranking reasoning {LLM}s under test-time scaling.
\newblock In {\em Proceedings of the 64th Annual Meeting of the Association for
  Computational Linguistics (Volume 1: Long Papers)}, pages 33437--33478.
  Association for Computational Linguistics.

\bibitem[Hariri et~al., 2025]{hariri2025dontpass}
Hariri, M., Samandar, A., Hinczewski, M., and Chaudhary, V. (2025).
\newblock Don't pass@k: A {Bayesian} framework for large language model
  evaluation.
\newblock arXiv preprint arXiv:2510.04265.

\bibitem[Hassid et~al., 2025]{hassid2025dontoverthink}
Hassid, M., Synnaeve, G., Adi, Y., and Schwartz, R. (2025).
\newblock Don't overthink it: Preferring shorter thinking chains for improved
  {LLM} reasoning.
\newblock arXiv preprint arXiv:2505.17813.

\bibitem[Hazon et~al., 2012]{hazon2012evaluation}
Hazon, N., Aumann, Y., Kraus, S., and Wooldridge, M. (2012).
\newblock On the evaluation of election outcomes under uncertainty.
\newblock {\em Artificial Intelligence}, 189:1--18.

\bibitem[Hendrycks et~al., 2021]{hendrycks2021math}
Hendrycks, D., Burns, C., Kadavath, S., Arora, A., Basart, S., Tang, E., Song,
  D., and Steinhardt, J. (2021).
\newblock Measuring mathematical problem solving with the {MATH} dataset.
\newblock In {\em Advances in Neural Information Processing Systems: Datasets
  and Benchmarks Track}.

\bibitem[Hoeffding, 1948]{hoeffding1948class}
Hoeffding, W. (1948).
\newblock A class of statistics with asymptotically normal distribution.
\newblock {\em The Annals of Mathematical Statistics}, 19(3):293--325.

\bibitem[Jana et~al., 2026]{jana2026proofbridge}
Jana, P., Kale, K., Tanriverdi, A., Song, C., Vishwanath, S., and Ganesh, V.
  (2026).
\newblock {ProofBridge}: Auto-formalization of natural language proofs in
  {Lean} via joint embeddings.
\newblock In {\em International Conference on Learning Representations}, pages
  13569--13591.

\bibitem[Kuncheva et~al., 2003]{kuncheva2003limits}
Kuncheva, L.~I., Whitaker, C.~J., Shipp, C.~A., and Duin, R. P.~W. (2003).
\newblock Limits on the majority vote accuracy in classifier fusion.
\newblock {\em Pattern Analysis and Applications}, 6(1):22--31.

\bibitem[Lehmann, 1983]{lehmann1983point}
Lehmann, E.~L. (1983).
\newblock {\em Theory of Point Estimation}.
\newblock Springer.

\bibitem[Li et~al., 2023]{li2023bird}
Li, J., Hui, B., Qu, G., Yang, J., Li, B., Li, B., Wang, B., Qin, B., Geng, R.,
  Huo, N., Zhou, X., Ma, C., Li, G., Chang, K., Huang, F., Cheng, R., and Li,
  Y. (2023).
\newblock Can {LLM} already serve as a database interface? a {BIg} bench for
  large-scale database grounded text-to-{SQL}s.
\newblock In {\em Advances in Neural Information Processing Systems},
  volume~36, pages 42330--42357. Curran Associates, Inc.

\bibitem[Li et~al., 2026]{li2026confidence}
Li, Y., Zhou, T., Shen, X., Miao, Y., Padman, R., and Krishnan, R. (2026).
\newblock Confidence as control: A survey of confidence utilization in large
  language models.
\newblock Preprint.

\bibitem[Liu et~al., 2025]{liu2025stable}
Liu, J., Liu, H., Xiao, L., Wang, Z., Liu, K., Gao, S., Zhang, W., Zhang, S.,
  and Chen, K. (2025).
\newblock Are your {LLM}s capable of stable reasoning?
\newblock In {\em Findings of the Association for Computational Linguistics:
  ACL 2025}, pages 17594--17632.

\bibitem[Liu, 2026]{liu2026twocalls}
Liu, Y. (2026).
\newblock Two calls, two moments, and the vote-accuracy curve of repeated {LLM}
  inference.
\newblock arXiv preprint arXiv:2605.03379.

\bibitem[{M-A-P Team} et~al., 2025]{mapteam2025supergpqa}
{M-A-P Team}, Du, X., Yao, Y., Ma, K., Wang, B., Zheng, T., Zhu, K., Liu, M.,
  Liang, Y., Jin, X., Wei, Z., Zheng, C., Deng, K., et~al. (2025).
\newblock {SuperGPQA}: Scaling {LLM} evaluation across 285 graduate
  disciplines.
\newblock arXiv preprint arXiv:2502.14739.

\bibitem[Marshall et~al., 2011]{marshall2011inequalities}
Marshall, A.~W., Olkin, I., and Arnold, B.~C. (2011).
\newblock {\em Inequalities: Theory of Majorization and Its Applications}.
\newblock Springer, second edition.

\bibitem[Nevius et~al., 1977]{nevius1977schur}
Nevius, S.~E., Proschan, F., and Sethuraman, J. (1977).
\newblock {Schur} functions in statistics {II}. {Stochastic} majorization.
\newblock {\em The Annals of Statistics}, 5(2):263--273.

\bibitem[Nowak, 2025]{nowak2025estimating}
Nowak, R. (2025).
\newblock Estimating the self-consistency of {LLM}s.
\newblock arXiv preprint arXiv:2509.19489.

\bibitem[Rein et~al., 2023]{rein2023gpqa}
Rein, D., Hou, B.~L., Stickland, A.~C., Petty, J., Pang, R.~Y., Dirani, J.,
  Michael, J., and Bowman, S.~R. (2023).
\newblock {GPQA}: A graduate-level {Google}-proof {Q\&A} benchmark.
\newblock arXiv preprint arXiv:2311.12022.

\bibitem[Saparkhan et~al., 2026]{saparkhan2026self}
Saparkhan, R., Hawasly, M., Parvez, M.~R., and Raza, M. (2026).
\newblock Self-consistency from only two samples: {CoT}--{PoT} ensembling for
  efficient {LLM} reasoning.
\newblock In {\em Findings of the Association for Computational Linguistics:
  ACL 2026}, pages 32804--32839. Association for Computational Linguistics.

\bibitem[Shah et~al., 2026]{shah2026upskill}
Shah, D., Yang, O., Yang, D., Zheng, C., and Eysenbach, B. (2026).
\newblock {UpSkill}: Mutual information skill learning for structured response
  diversity in {LLM}s.
\newblock arXiv preprint arXiv:2602.22296.

\bibitem[Stoye, 2009]{stoye2009partial}
Stoye, J. (2009).
\newblock Partial identification and robust treatment choice: An application to
  young offenders.
\newblock {\em Journal of Statistical Theory and Practice}, 3(1):239--254.

\bibitem[Suzgun et~al., 2023]{suzgun2023challenging}
Suzgun, M., Scales, N., Sch{\"a}rli, N., Gehrmann, S., Tay, Y., Chung, H.~W.,
  Chowdhery, A., Le, Q.~V., Chi, E.~H., Zhou, D., and Wei, J. (2023).
\newblock Challenging {BIG}-bench tasks and whether chain-of-thought can solve
  them.
\newblock In {\em Findings of the Association for Computational Linguistics:
  ACL 2023}, pages 13003--13051.

\bibitem[Tamer, 2010]{tamer2010partial}
Tamer, E. (2010).
\newblock Partial identification in econometrics.
\newblock {\em Annual Review of Economics}, 2(1):167--195.

\bibitem[Wang et~al., 2023]{wang2023selfconsistency}
Wang, X., Wei, J., Schuurmans, D., Le, Q.~V., Chi, E.~H., Narang, S.,
  Chowdhery, A., and Zhou, D. (2023).
\newblock Self-consistency improves chain of thought reasoning in language
  models.
\newblock In {\em International Conference on Learning Representations}.

\bibitem[Wunderlich et~al., 2026]{wunderlich2026multi}
Wunderlich, F.~V., Kaesberg, L.~B., Wahle, J.~P., Ruas, T., and Gipp, B.
  (2026).
\newblock Multi-agent reasoning improves compute efficiency: {Pareto}-optimal
  test-time scaling.
\newblock In {\em Proceedings of the 64th Annual Meeting of the Association for
  Computational Linguistics (Volume 4: Student Research Workshop)}, pages
  1--14. Association for Computational Linguistics.

\bibitem[Xia et~al., 2024]{xia2024sportqa}
Xia, H., Yang, Z., Wang, Y., Tracy, R., Zhao, Y., Huang, D., Chen, Z., Zhu, Y.,
  Wang, Y.-f., and Shen, W. (2024).
\newblock {SportQA}: A benchmark for sports understanding in large language
  models.
\newblock In {\em Proceedings of the 2024 Conference of the North American
  Chapter of the Association for Computational Linguistics: Human Language
  Technologies (Volume 1: Long Papers)}, pages 5061--5081. Association for
  Computational Linguistics.

\bibitem[Yang et~al., 2026]{yang2026correctmass}
Yang, H., Sun, J., and Ma, X. (2026).
\newblock More correct mass, worse answers: Why power sampling can fail and how
  to fix it.
\newblock arXiv preprint arXiv:2608.14420.

\bibitem[Yu et~al., 2018]{yu2018spider}
Yu, T., Zhang, R., Yang, K., Yasunaga, M., Wang, D., Li, Z., Ma, J., Li, I.,
  Yao, Q., Roman, S., Zhang, Z., and Radev, D. (2018).
\newblock {Spider}: A large-scale human-labeled dataset for complex and
  cross-domain semantic parsing and text-to-{SQL} task.
\newblock In {\em Proceedings of the 2018 Conference on Empirical Methods in
  Natural Language Processing}, pages 3911--3921. Association for Computational
  Linguistics.

\bibitem[Zhang et~al., 2026]{zhang2026decomposing}
Zhang, L., Tang, M., Long, C., Tang, X., and Luo, X. (2026).
\newblock Decomposing wrong-consensus agreement in {LLM} self-consistency.
\newblock arXiv preprint arXiv:2608.18795.

\end{thebibliography}

\clearpage
\appendix
\thispagestyle{empty}
\onecolumn
\raggedbottom
\aistatstitle{Self-Consistency from Incomplete Logs: Supplementary Materials}
\setlength{\belowcaptionskip}{4pt}

\section{Proofs}
\label{app:proofs}

\subsection{Proof of Theorem~\ref{thm:bounds}}

Fix the physical pool: $c$ gold outputs and $n-c$ individually labeled wrong
outputs.  An \emph{allocation} is a partition of the wrong outputs into answer
categories.  Couple all allocations by drawing the same uniform physical
$k$-subset.

For a realized subset, let $r$ be its number of gold outputs and let
$w=(w_1,w_2,\ldots)$ be its wrong-category counts.  Its tie-adjusted gold payoff
is zero when $r=0$ and otherwise is
\[
g(r,w)=\frac{\mathbf{1}\{r\geq\max_j w_j\}}
{1+\sum_j\mathbf{1}\{w_j=r\}}.
\]
Here the maximum over an empty wrong-category vector is zero.
If wrong categories of selected sizes $x,y$ are merged and $x+y>r$, the
resulting payoff is zero. If $x+y=r$, merging two positive counts creates a
new opponent tied with gold; merging a zero and a positive count leaves the
old tie unchanged. If $x+y<r$, neither count affects the top tie before or
after merging. Thus the gold payoff cannot increase. Equivalently, splitting
one wrong category cannot decrease
$g(r,w)$.  Iterating mergers reaches one wrong category and iterating splits
reaches singleton wrong categories.  Taking expectations under the common
physical-subset draw preserves both inequalities.

With one wrong category, gold wins if $r>k-r$, ties one opponent if $r=k-r$,
and loses otherwise, yielding Eq.~\eqref{eq:lower}.  With singleton wrong
categories, gold wins uniquely for $r\geq2$; for $r=1$, all $k$ selected
answers tie and gold receives $1/k$; for $r=0$ it loses.  This gives
Eq.~\eqref{eq:upper}.  Both allocations exist for every $n$ and $c$, hence the
bounds are sharp. \hfill$\square$

\subsection{Pairwise smoothing and proof of Theorem~\ref{thm:boundedq}}
\label{app:smoothing}

\paragraph{Classical reduction.}
Conditional on $r>0$ selected gold outputs, the wrong-count vector $X$ is
multivariate hypergeometric with population counts $w$ and sample size $k-r$.
The payoff $f_r(x)=\mathbf{1}\{\max_jx_j\leq r\}/(1+\#\{j:x_j=r\})$
is symmetric and Schur-concave on the integer simplex. A balancing transfer
cannot lower it: while a count exceeds $r$ the payoff is zero; once all counts
are at most $r$, balancing cannot create a new tie at $r$ and can remove one.
Hypergeometric expectation preserves this ordering
\citep[Proposition~11.E.14]{marshall2011inequalities}; averaging over $r$ uses
weights independent of $w$. Thus $V_k(c,w)$ is Schur-concave in $w$, with the
balanced and concentrated extrema. Degenerate conditional samples are
immediate. The elementary proof below reaches the same conclusion without the
general preservation theorem.

\begin{lemma}[One-output smoothing]
Fix the gold count, all other category sizes and the vote budget. If two
wrong categories have sizes $a\geq b+2$, replacing $(a,b)$ by $(a-1,b+1)$
cannot decrease the uniform-subset, uniform-tie gold payoff.
\label{lem:smoothing}
\end{lemma}

Label one physical output $z$ in the larger category and move only its category
label. Couple both allocations on the same physical subset. Subsets without
$z$ have unchanged payoff. For subsets containing $z$, condition on all
selected outputs outside these two categories. Let $r$ be the selected gold
count and $s$ the selected total from the pair, including $z$. If $r=0$ or
another wrong category exceeds $r$, both payoffs are zero. Otherwise let $T$
be the number of other wrong categories tied at $r$ and define
\[
\phi(x,y)=\frac{\mathbf{1}\{x\leq r,y\leq r\}}
{1+T+\mathbf{1}\{x=r\}+\mathbf{1}\{y=r\}}.
\]
For $x+y=s$, the following exhaustive cases show why this function is symmetric
and nondecreasing toward the balanced pair. Write $M=\max(x,y)$; balancing
decreases $M$. Infeasible rows for a particular integer $s$ are omitted.
\begin{center}
\begin{tabular}{lll}
\toprule
Fixed pair total & Pair condition & Payoff $\phi(x,y)$ \\
\midrule
$s>2r$ & Every pair & $0$ \\
$s=2r$ & $(x,y)\ne(r,r)$ & $0$ \\
$s=2r$ & $(x,y)=(r,r)$ & $1/(T+3)$ \\
$s<2r$ & $M>r$ & $0$ \\
$s<2r$ & $M=r$ (one tied opponent) & $1/(T+2)$ \\
$s<2r$ & $M<r$ & $1/(T+1)$ \\
\bottomrule
\end{tabular}
\end{center}

Write $A=a-1$, $B=b$. A remaining pair selection $(u,v)$ with $u+v=s-1$ has
weight $W(u,v)=\binom Au\binom Bv$ and payoff change
\[
\Delta(u,v)=\phi(u,v+1)-\phi(u+1,v).
\]
Symmetry gives $\Delta(v,u)=-\Delta(u,v)$; balancing gives
$\Delta(u,v)\geq0$ for $u>v$. Moreover $W(u,v)\geq W(v,u)$ for $u>v$:
when $u\leq B$, the ratio $\binom Aj/\binom Bj$ is nondecreasing in $j$
because its successive ratio is $(A-j)/(B-j)\geq1$; when $u>B$, the
right-hand weight is zero. Out-of-support binomial coefficients are zero.
Pairing the terms leaves
\[
\sum_{u>v,\ u+v=s-1}
\bigl[W(u,v)-W(v,u)\bigr]\Delta(u,v)\geq0,
\]
with a common unchanged denominator $\binom{A+B}{s-1}$. This proves the
conditional inequality and hence the lemma after averaging.

To prove Theorem~\ref{thm:boundedq}, pad any allocation to length $q$ with
zeros and repeatedly smooth pairs differing by at least two. Each operation
strictly decreases the integer sum of squared counts, so the process ends
at $b_q(n-c)$. The lemma proves the upper inequality and the balanced vector
attains it. The lower endpoint follows from the merging argument for
Theorem~\ref{thm:bounds}, whose concentrated vector is feasible for every
$q\geq1$. \hfill$\square$

\subsection{Proof of Corollary~\ref{cor:abstention}}

Conditioning on the number $v$ of selected valid outputs gives weight $w_v$
and a uniform subset of the valid pool. Apply Theorem~\ref{thm:boundedq}
conditionally and average. Concentrated and balanced allocations attain the
endpoints for all $v$ simultaneously, which establishes sharpness. The
unrestricted case uses singleton wrong categories.

If $a$ is unknown, replace each abstention by a wrong vote. This cannot
increase gold payoff: it adds competitors but no correct votes. Merging
all wrong categories then reaches the lower construction with $a=0$.
Conversely, no subset without gold can succeed; making every noncorrect
output abstain makes every subset containing gold succeed. These two
constructions attain the stated endpoints. \hfill$\square$

\subsection{Proof of Theorem~\ref{thm:ustat}}

For every fixed $S\subset[n]$ of size $k$, $(X_i)_{i\in S}\sim p^k$; hence
$\E[h_k(X_S)]=\theta_k(p)$.  Averaging these equal expectations over all
$\binom nk$ subsets proves $\E[U_{n,k}]=\theta_k(p)$.

Now independently draw $S$ uniformly from the size-$k$ subsets and set
$T=h_k(X_S)$.  Conditional on the stored sample,
\[
\E[T\mid X_1,\ldots,X_n]
=\binom nk^{-1}\sum_{|S|=k}h_k(X_S)=U_{n,k}.
\]
The conditional-variance identity gives
\[
\operatorname{Var}(T)=
\operatorname{Var}(U_{n,k})+
\E[\operatorname{Var}(T\mid X_1,\ldots,X_n)],
\]
which proves the variance inequality.  For a fixed finite alphabet and
$p$ in the interior of its simplex, $C$ is a complete sufficient statistic for
the multinomial family.  Because $U_{n,k}$ is a function of $C$ and is unbiased,
the Lehmann--Scheff\'{e} theorem gives the unique uniformly minimum-variance
unbiased estimator in this fixed-alphabet family \citep{lehmann1983point}.
\hfill$\square$

\subsection{Proof of Proposition~\ref{prop:contrast}}
\label{app:contrastproof}

Every compatible physical pool assigns its $e$ wrong outputs to at most $q$
categories. Sorting and padding these category counts gives an element of
$\mathcal W_{e,q}$ without changing any uniformly tie-broken payoff. Conversely,
every $w\in\mathcal W_{e,q}$ can be realized by assigning the $e$ physical
wrong outputs to categories of those sizes, alongside the fixed $c$ correct
outputs and $a$ abstentions. Therefore the finite set
\[
\{F_\ell(w)-F_k(w):w\in\mathcal W_{e,q}\}
\]
is exactly the identified set for the gain, and its minimum and maximum are
attainable. Because $L_j\leq F_j(w)\leq B_j$ for every $w$ and every budget
$j$, each gain lies in $[L_\ell-B_k,B_\ell-L_k]$. The two single-budget
endpoints need not be attained by the same $w$.

For items $i=1,\ldots,M$, assume that their compatible allocation set is the
Cartesian product $\prod_i\mathcal W_{e_i,q_i}$. Then
\[
\min_{(w_i)}\frac1M\sum_i\Delta_i(w_i)
 =\frac1M\sum_i\min_{w_i}\Delta_i(w_i),
\]
and similarly for the maximum: select a minimizing or maximizing witness
independently for each item. This step uses no independence assumption on the
generations. The endpoints are sharp, but not every value between them need be
attainable.
\hfill$\square$

\section{Dynamic programs}
\label{app:algorithm}

\subsection{Correctness and complexity of the count recurrence}

Fix $r\geq1$.  We prove by induction on $j$ that $D_j(s,t)$ in
Eq.~\eqref{eq:dp} equals the number of physical ways to select $s$ outputs from
the first $j$ wrong categories while selecting at most $r$ from each and
selecting exactly $r$ from exactly $t$ categories.  The claim is immediate for
$j=0$.  For category $j$, selecting $\ell$ of its $C_j$ physical outputs has
$\binom{C_j}{\ell}$ possibilities, increments $s$ by $\ell$, and increments
$t$ exactly when $\ell=r$.  Summing over all feasible $\ell$ proves the
induction step.

For total selected gold count $r$, there are $\binom{C_0}{r}$ ways to choose
the physical gold outputs.  A DP terminal state $D_m(k-r,t)$ contains no wrong
count exceeding $r$, so gold is a plurality winner tied with exactly $t$ wrong
categories and receives payoff $1/(t+1)$.  Thus Eq.~\eqref{eq:numerator}
counts the sum of the kernel over every physical $k$-subset exactly once.

The implementation stores only reachable $(s,t)$ pairs.  For each $r$, it
processes $m$ categories, $s\leq k-r$, $t\leq k-r$, and at most $r+1$ values of
$\ell$ per state, so a loose worst-case bound for the displayed recurrence is
$O(mk^4)$ time after summing over $r$ and $O(k^2)$ working states per layer.
The practical state space is much smaller.  Integer binomial weights preserve
combinatorial exactness; only tie fractions and the final division use
floating-point arithmetic. On small pools, including binary answers,
abstentions and several correct categories, the implementation matches
brute-force enumeration to within $10^{-12}$.

\subsection{All-budget recurrence}
\label{app:allbudget}

For fixed selected correct count $r>0$, let $z$ track selected wrong outputs
and $y$ track categories tied with the correct count. For a wrong category
with $w_j$ stored outputs, its non-losing factor is
\[
 A_{j,r}(z,y)=\sum_{s=0}^{\min(w_j,r-1)}\binom{w_j}{s}z^s
       +\mathbf1\{w_j\geq r\}\binom{w_j}{r}z^ry.
\]
Let $H_r(z,y)=\prod_j A_{j,r}(z,y)$. Replacing a coefficient of $y^t$ by
its weight $1/(1+t)$ integrates uniform tie breaking. Abstentions contribute
$(1+z)^a$ but never a vote. Hence, for each $k\geq1$,
\[
 F_k(w)=\frac{1}{\binom nk}\sum_{r=1}^{\min(c,k)}\binom cr
    [z^{k-r}](1+z)^a\sum_{t=0}^{q}\frac{[y^t]H_r(z,y)}{1+t}.
\]
The zero-correct and all-abstention outcomes contribute zero. A single set
of polynomial products truncated at degree $K=\max\mathcal K$ supplies
every requested budget $k\in\mathcal K$. The implemented dense recurrence,
including convolution with the abstention polynomial, uses
$O(q^2K^3+nK^2)$ arithmetic operations and $O(qK+n)$ numerical working
space per allocation, apart from preparing cached binomial coefficients.
It also retains $O(|\mathcal W_{e,q}|(q+|\mathcal K|))$ allocation and curve
entries when taking extrema. The implementation uses
double precision. Enumeration is polynomial in $e$ at fixed $q$; the displayed bound
on allocation count is not polynomial uniformly in both $e$ and $q$.

\section{Data and answer processing}
\label{app:data}

\subsection{Scorio Lite and Trace}
\label{app:scorio}

We use fixed revisions of Scorio Lite, Scorio Trace and the SuperGPQA catalog
of answer options. Their revision identifiers are recorded in the supplement. The Lite
release contributes 59,520 mathematics attempts and 1,152,000 SuperGPQA
attempts, and the Trace release contributes 192,000 mathematics attempts. Each
model--question pair has 80 outputs, which we call a pool. In 287 pools, the
correct outputs fall into more than one answer category; Plurality-U evaluates
these pools directly, but the identification analysis assumes a single correct
category and therefore omits them.

The SuperGPQA questions have four to ten options, and 3,203 of the 3,600 have
ten.

The data are available at
\url{https://huggingface.co/datasets/harimo/scorio-lite},
\url{https://huggingface.co/datasets/harimo/scorio-trace} and
\url{https://huggingface.co/datasets/m-a-p/SuperGPQA}. The two Scorio dataset
cards declare the MIT license, and the SuperGPQA repository declares ODC-By and
notes that incorporated source questions keep their original terms. We release
the derived answer histograms, not prompts or generated text.

\subsection{MATH and GPQA pools}

The MATH pools come from
\url{https://huggingface.co/datasets/ChrisMcCormick/math-rollouts} and contain
outputs of Qwen2.5-Math-1.5B on MATH-500 and of its Oat-Zero variant on
MATH-500 and on 716 other MATH problems. The GPQA pools come from the public
artifact accompanying \citet{bahuguna2026backfires} and contain outputs of
\texttt{Qwen2.5-7B-Instruct-Turbo} and
\texttt{Meta-Llama-3-8B-Instruct-Lite} on the 198 GPQA Diamond questions.
Every pool stores 64 outputs per question. The math-rollouts dataset card
declares the MIT license, and the GPQA rollout repository declares Apache-2.0.
Our released derivatives contain aggregate statistics and normalized answer
categories.

The math outputs were generated with temperature 0.6 and top-$p$ 0.95. None
of the sources used in this paper records random seeds, so the independence of
generations cannot be checked from the released logs.

\subsection{Answer extraction and category rules}
\label{app:normalizers}

Before voting, each output is mapped to an answer category. For GPQA, the
category is the answer letter, and an output without a valid letter abstains
and casts no vote. For math, the category is the first mathematical object that
the Math-Verify parser (\texttt{math\_verify}) extracts from the output. Math outputs that
fail to parse form one shared failure category, which receives votes like any
other category. Correctness labels play no part in this grouping. After the
vote, we score the result by drawing one output at random from the winning
category and reading its released correctness label. In the retained data, no
category contains both correct and incorrect outputs, but the correct outputs
of a question can fall into several categories: this happens in 123/500, 6/500
and 12/716 questions of the three math pools.

Table~\ref{tab:policysensitivity} compares two alternative rules: retaining
only the last boxed answer of each math output, and treating invalid outputs
as one shared wrong answer instead of abstentions. The latter has little
effect on SuperGPQA and GPQA but a large effect on Adaptive-Consistency
(Section~\ref{sec:exp-policy}).

\begin{table}[H]
\caption{Sensitivity to answer processing at $k=9$ (pp). Change is
alternative-rule accuracy minus our-rule accuracy for base MATH-500
(Qwen2.5-Math-1.5B); the other rows give maximum absolute changes across
model--task configurations.}
\label{tab:policysensitivity}
\centering\footnotesize
\setlength{\tabcolsep}{2.5pt}
\begin{tabular}{@{}lp{6.4cm}r@{}}
\toprule
Dataset & Alternative rule & Change \\
\midrule
MATH-500 (base) & Keep only the last boxed answer & $-17.56$ \\
SuperGPQA & Count invalid outputs as one wrong answer & $\leq0.25$ \\
GPQA Diamond & Count invalid outputs as one wrong answer & $\leq0.30$ \\
Adaptive-Consistency & Count invalid outputs as one wrong answer & $\leq49.11$ \\
\bottomrule
\end{tabular}
\end{table}

\begin{table}[H]
\caption{Math voting accuracy at $k=9$ under three answer-processing rules
(\%). Math-Verify (our rule) and Last box group answers without correctness
labels; Oracle merge uses the labels and serves only as a reference. Box
failure is the percentage of outputs with no usable last box. Base denotes
Qwen2.5-Math-1.5B; Oat-Zero is its Oat-Zero variant, and Other MATH contains
the 716 problems outside MATH-500.}
\label{tab:normalizers}
\centering
\small
\begin{tabular}{lrrrr}
\toprule
Pool & Oracle merge & Math-Verify & Last box & Box failure (\%) \\
\midrule
Base / MATH-500 & 55.76 & 54.92 & 37.36 & 36.50 \\
Oat-Zero / MATH-500 & 78.57 & 78.53 & 78.41 & 1.28 \\
Oat-Zero / Other MATH & 52.73 & 52.67 & 52.19 & 3.04 \\
\bottomrule
\end{tabular}
\end{table}

Table~\ref{tab:normalizers} compares three math rules and reports last-box
extraction failures. The last-box
rule takes the last \texttt{boxed} or \texttt{fbox} answer as written, removing
only whitespace and applying no algebraic simplification; outputs without a
usable box form one failure category, which scores zero if it wins. On base
MATH, this rule lowers accuracy from 54.92\% to 37.36\%, whereas on the two
Oat-Zero pools it changes accuracy by less than half a point. The oracle rule
merges all correct answers of a question into one category using the
correctness labels, which raises the $k=9$ scores by 0.84, 0.04 and 0.06
points. A deployed selector has no access to these labels, so we use the
oracle rule only as a reference.

For example, on base-MATH question algebra/7515, four correct outputs give
$11/2$ and nine give the decimal value $5.5$. The source labels both forms correct,
but our rule places them in separate answer categories, which splits their
votes.

\section{Analysis protocol and timing study}
\label{app:protocol}

\subsection{Protocol}

The supplement contains the data, metrics, analysis plans and reproduction
code. Each plan was written down before the outcomes of that analysis were
inspected, without external registration. The math answer rule of
Appendix~\ref{app:normalizers} was chosen after the initial results.

\subsection{Timing study}
\label{app:operational}

The timing study uses the answer-processing rules of
Appendix~\ref{app:normalizers}.

\begin{figure}[!htbp]
\centering
\includegraphics[width=0.96\textwidth]{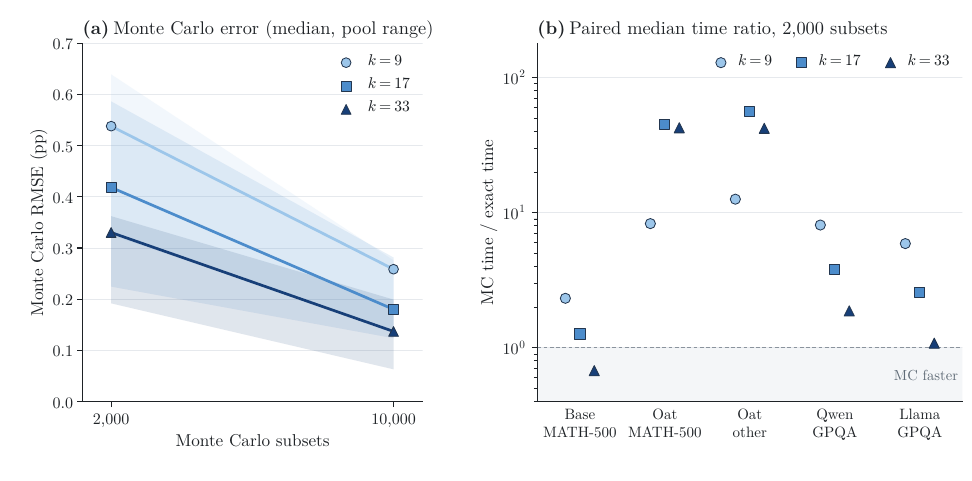}
\caption{\textbf{Exact versus Monte Carlo computation.} Median Monte Carlo
RMSE over the five timing pools (left; shading gives the pool range) and paired
median Monte Carlo/exact runtime ratios at 2,000 subsets (right). Ratios above
one favor exact computation; the only cell below one is base MATH at $k=33$.}
\label{fig:runtime}
\end{figure}

Table~\ref{tab:runtimecells} reports runtime ratios for each pool and vote budget, and
Figure~\ref{fig:runtime} shows the Monte Carlo error and the runtime ratios.

\begin{table}[H]
\caption{Monte Carlo/exact runtime ratios: for each question, the median Monte
Carlo time divided by the median exact time, and then the median over the ten
questions. Each entry reports $B=2{,}000\,/\,10{,}000$ subsets; values above
one favor exact computation. Math pool names follow
Table~\ref{tab:normalizers}; Llama and Qwen denote the Llama-3-8B and
Qwen2.5-7B pools of Table~\ref{tab:data}.}
\label{tab:runtimecells}
\centering\footnotesize
\setlength{\tabcolsep}{2.5pt}
\begin{tabular}{@{}lrrr@{}}
\toprule
Pool & $k=9$ & $k=17$ & $k=33$ \\
\midrule
Base / MATH-500 & 2.33/11.63 & 1.27/6.25 & 0.68/3.33 \\
Oat-Zero / MATH-500 & 8.32/30.65 & 45.35/215.94 & 42.63/202.71 \\
Oat-Zero / Other MATH & 12.61/49.99 & 56.15/270.00 & 42.24/199.26 \\
Llama / GPQA Diamond & 5.92/27.30 & 2.59/12.34 & 1.08/5.30 \\
Qwen / GPQA Diamond & 8.12/38.13 & 3.80/18.33 & 1.88/8.94 \\
\bottomrule
\end{tabular}
\end{table}

Exact evaluation becomes slower as answers spread across more categories.
Base MATH, the only pool where Monte Carlo is sometimes faster, has a median
of 22 answer categories per question, compared with at most 5.5 in the other
pools.

\section{Adaptive-Consistency pools}
\label{app:independent}

\paragraph{Selection.}
We use the public outputs released with Adaptive-Consistency
\citep{aggarwal2023adaptive}, with 40 generations per question from
code-davinci-002 and Vicuna-13B. We use every task except the numeric-answer
tasks \texttt{gsm}, \texttt{svamp} and \texttt{asdiv}: nine BIG-Bench Hard
tasks \citep{suzgun2023challenging} and StrategyQA \citep{geva2021did}.
Logical Deduction and Tracking Shuffled Objects are the three-object variants
used by the source. We exclude 64 unusable rows (targets outside the displayed options, truncated
option lists, or only one generation), which leaves 8,880 model--question
units with 4,469 distinct questions.

\paragraph{Answer processing.}
The source provides extracted answers, which we match to the displayed
options (\texttt{True}/\texttt{False} for Boolean Expressions and
\texttt{yes}/\texttt{no} for StrategyQA). Date Understanding has no fixed options,
so any nonempty answer forms a category; it is the only task without a
fixed option set. Invalid outputs abstain, and a subset with only invalid
outputs scores zero. Ties are broken uniformly at random, which can differ
from the source's own adaptive evaluation. The source reports temperature 0.7
and top-$p$ 1.

\paragraph{Uncertainty.}
Table~\ref{tab:independentaccuracy} gives paired bootstrap 95\% intervals over
questions. All twenty gaps remain positive after a Bonferroni correction
across 60 metric contrasts (three per pool).

\begin{table}[!htbp]
\caption{All Adaptive-Consistency pools at $k=9$ (accuracy in \%). Avg@1
is mean single-response correctness; Strict maj. is strict-majority
correctness. Gap is Plurality-U minus Strict maj. (pp), with paired bootstrap
95\% confidence intervals over questions.}
\label{tab:independentaccuracy}
\centering\small
\setlength{\tabcolsep}{5pt}
\begin{tabular}{llrrrrr}
\toprule
Model & Task & Questions & Avg@1 & Strict maj. & Plurality-U & Gap [95\% CI] \\
\midrule
code-davinci-002 & Boolean Expressions & 250 & 86.72 & 92.74 & 93.57 & 0.83 [0.46, 1.30] \\
code-davinci-002 & Date Understanding & 369 & 67.28 & 72.10 & 77.64 & 5.54 [4.33, 6.85] \\
code-davinci-002 & Disambiguation QA & 250 & 61.15 & 66.38 & 73.71 & 7.33 [5.84, 8.98] \\
code-davinci-002 & Logical Deduction & 250 & 81.75 & 87.13 & 88.73 & 1.60 [1.06, 2.19] \\
code-davinci-002 & Penguins & 146 & 73.46 & 78.02 & 82.20 & 4.18 [2.95, 5.54] \\
code-davinci-002 & Ruin Names & 248 & 49.65 & 51.42 & 76.63 & 25.20 [22.31, 28.21] \\
code-davinci-002 & Salient Translation & 250 & 56.49 & 58.86 & 63.12 & 4.26 [3.02, 5.63] \\
code-davinci-002 & Snarks & 177 & 63.73 & 68.13 & 74.12 & 5.99 [4.70, 7.41] \\
code-davinci-002 & StrategyQA & 2,279 & 71.15 & 77.22 & 77.69 & 0.48 [0.42, 0.54] \\
code-davinci-002 & Tracking Shuffled Objects & 250 & 71.07 & 74.28 & 76.02 & 1.74 [1.23, 2.30] \\
\addlinespace
Vicuna-13B & Boolean Expressions & 250 & 42.25 & 37.54 & 75.87 & 38.33 [35.15, 41.51] \\
Vicuna-13B & Date Understanding & 369 & 39.49 & 40.83 & 57.00 & 16.17 [13.86, 18.56] \\
Vicuna-13B & Disambiguation QA & 250 & 31.00 & 20.61 & 58.92 & 38.32 [35.13, 41.49] \\
Vicuna-13B & Logical Deduction & 250 & 28.92 & 23.47 & 48.93 & 25.47 [22.28, 28.69] \\
Vicuna-13B & Penguins & 146 & 24.38 & 17.35 & 43.87 & 26.52 [22.53, 30.56] \\
Vicuna-13B & Ruin Names & 248 & 14.76 & 3.16 & 39.40 & 36.24 [32.88, 39.64] \\
Vicuna-13B & Salient Translation & 192 & 14.99 & 4.60 & 28.31 & 23.71 [20.44, 27.07] \\
Vicuna-13B & Snarks & 177 & 28.18 & 15.49 & 68.29 & 52.80 [48.85, 56.66] \\
Vicuna-13B & StrategyQA & 2,279 & 37.45 & 34.82 & 64.47 & 29.65 [28.51, 30.78] \\
Vicuna-13B & Tracking Shuffled Objects & 250 & 23.73 & 16.18 & 32.45 & 16.27 [13.91, 18.72] \\
\bottomrule

\end{tabular}
\end{table}

\paragraph{Results.}
Table~\ref{tab:independentaccuracy} reports all twenty pools, and
Table~\ref{tab:independentlogging} reports the interval widths under each log
and the sensitivity to the rule for invalid outputs. The six binary pools have
zero width once $a$ and $q$ are known. With no abstentions and odd $k$, exact
voting accuracy equals strict-majority correctness, as Section~4 predicts.
With abstentions, this identity holds for the valid votes rather than for all
slots. The gap in Table~\ref{tab:independentaccuracy} comes from the change of
scoring rule, whereas the interval width comes from missing answer-category
information.

\paragraph{Source rights.}
The Adaptive-Consistency code repository declares Apache-2.0, and we could not
confirm a separate license for its externally hosted generation archive. We
therefore release only derived answer histograms, not raw generations or
question text. All results in this appendix can be reproduced from these
histograms.

\begin{table}[!htbp]
\caption{Adaptive-Consistency at $k=9$. The interval-width columns follow
Table~\ref{tab:abstentionbounds}: correctness labels only, then abstention
counts ($+a$), then the option count ($+a,q$). The last two columns treat all
invalid outputs of a question as one incorrect answer instead of abstentions:
Accuracy is the resulting voting accuracy, and $\Delta$ is our accuracy minus
this accuracy. Date Understanding has no fixed options, so its $+a$ and $+a,q$
widths coincide.}
\label{tab:independentlogging}
\centering\small
\begin{tabular}{llrrrrr}
\toprule
 & & \multicolumn{3}{c}{Interval width (pp)} & \multicolumn{2}{c}{Invalid as one answer} \\
\cmidrule(lr){3-5}\cmidrule(l){6-7}
Model & Task & Labels & $+a$ & $+a,q$ & Accuracy (\%) & $\Delta$ (pp) \\
\midrule
code-davinci-002 & Boolean Expressions & 6.53 & 4.73 & 0.00 & 93.50 & 0.07 \\
code-davinci-002 & Date Understanding & 16.78 & 13.43 & 13.43 & 77.62 & 0.02 \\
code-davinci-002 & Disambiguation QA & 28.10 & 18.84 & 6.60 & 72.28 & 1.43 \\
code-davinci-002 & Logical Deduction & 11.20 & 8.80 & 3.11 & 88.72 & 0.01 \\
code-davinci-002 & Penguins & 19.57 & 15.20 & 9.67 & 82.13 & 0.07 \\
code-davinci-002 & Ruin Names & 42.88 & 14.26 & 7.67 & 61.39 & 15.23 \\
code-davinci-002 & Salient Translation & 22.26 & 14.00 & 8.27 & 62.54 & 0.57 \\
code-davinci-002 & Snarks & 28.99 & 18.04 & 0.00 & 72.88 & 1.24 \\
code-davinci-002 & StrategyQA & 20.89 & 17.62 & 0.00 & 77.69 & 0.00 \\
code-davinci-002 & Tracking Shuffled Objects & 18.80 & 13.85 & 4.28 & 76.01 & 0.01 \\
\addlinespace
Vicuna-13B & Boolean Expressions & 56.52 & 11.87 & 0.00 & 45.41 & 30.46 \\
Vicuna-13B & Date Understanding & 29.70 & 11.49 & 11.49 & 48.04 & 8.96 \\
Vicuna-13B & Disambiguation QA & 65.44 & 18.39 & 6.65 & 30.28 & 28.64 \\
Vicuna-13B & Logical Deduction & 50.25 & 15.30 & 4.94 & 30.68 & 18.25 \\
Vicuna-13B & Penguins & 52.65 & 16.62 & 10.23 & 24.54 & 19.33 \\
Vicuna-13B & Ruin Names & 61.77 & 13.66 & 7.68 & 6.62 & 32.78 \\
Vicuna-13B & Salient Translation & 55.86 & 21.46 & 14.28 & 12.34 & 15.97 \\
Vicuna-13B & Snarks & 70.43 & 8.35 & 0.00 & 19.18 & 49.11 \\
Vicuna-13B & StrategyQA & 47.86 & 10.52 & 0.00 & 39.67 & 24.81 \\
Vicuna-13B & Tracking Shuffled Objects & 52.62 & 24.71 & 6.96 & 24.75 & 7.70 \\
\bottomrule

\end{tabular}
\end{table}

\FloatBarrier
\section{Budget-contrast details}
\label{app:budget}

\subsection{Data}

The eligible records and the three budget pairs were fixed after earlier
analyses of the same source data and before the budget contrasts were
computed. The comparison includes every GPQA and Adaptive-Consistency question
with a fixed option set. It excludes MATH, whose correct answers can fall into
several categories, and Date Understanding, which has no fixed options. The
intervals are deterministic identification intervals, not bootstrap
confidence intervals.

The SuperGPQA comparison uses all 3,600 questions of each of the four Scorio
Lite models, or 14,400 model--question pairs. Its plan was written after the
comparisons on the other pools and before any SuperGPQA gain was computed, and
it uses the same budget pairs and evaluator. Pairs with the same $(n,c,a,q)$
share an allocation set, so the 1,445 distinct profiles require 90,347,322
allocation evaluations. Every question-level interval contains the gain
computed from the full log.

\subsection{Results for each pool}

Tables~\ref{tab:allbudgetpools} and~\ref{tab:budgetsuper} list the separate
and shared intervals of every pool with more than two options,
and Table~\ref{tab:budgetsummary} summarizes all three budget comparisons.

\begin{table}[H]
\caption{Pool-average accuracy changes from 9 to 17 votes in all 14 pools
with more than two options (pp). Separate combines bounds optimized for each
budget; Shared bounds their difference over the same answer allocation.
Both use the same $(n,c,a,q)$. Exact change is computed from full answer
counts.}
\label{tab:allbudgetpools}
\centering\footnotesize
\begin{tabular}{@{}llrrrr@{}}
\toprule
Model & Task & Questions & Separate & Shared & Exact change \\
\midrule
code-davinci-002 & Disambiguation QA & 250 & $[-5.76,9.20]$ & $[-0.33,3.64]$ & 1.58 \\
code-davinci-002 & Logical Deduction & 250 & $[-2.77,4.21]$ & $[-0.21,1.59]$ & 0.53 \\
code-davinci-002 & Penguins & 146 & $[-8.65,12.26]$ & $[-0.61,4.46]$ & 1.22 \\
code-davinci-002 & Ruin Names & 248 & $[-6.71,11.46]$ & $[-0.39,5.06]$ & 1.61 \\
code-davinci-002 & Salient Translation & 250 & $[-8.07,10.31]$ & $[-1.35,3.50]$ & 0.77 \\
code-davinci-002 & Tracking Shuffled Objects & 250 & $[-3.91,5.41]$ & $[-0.49,1.83]$ & 0.42 \\
Llama-3-8B & GPQA Diamond & 198 & $[-14.00,16.04]$ & $[-2.52,3.98]$ & 0.65 \\
Qwen2.5-7B & GPQA Diamond & 198 & $[-12.58,14.29]$ & $[-2.46,3.74]$ & 0.32 \\
\midrule
Vicuna-13B & Disambiguation QA & 250 & $[-5.58,11.04]$ & $[0.24,5.12]$ & 2.30 \\
Vicuna-13B & Logical Deduction & 250 & $[-4.78,7.12]$ & $[-0.65,2.90]$ & 0.89 \\
Vicuna-13B & Penguins & 146 & $[-11.21,15.32]$ & $[-2.31,6.22]$ & 1.00 \\
Vicuna-13B & Ruin Names & 248 & $[-8.29,14.89]$ & $[-1.12,7.63]$ & 3.33 \\
Vicuna-13B & Salient Translation & 192 & $[-16.28,21.49]$ & $[-3.42,8.26]$ & 1.80 \\
Vicuna-13B & Tracking Shuffled Objects & 250 & $[-8.27,8.70]$ & $[-2.34,2.57]$ & $-0.22$ \\
\bottomrule
\end{tabular}
\end{table}

\begin{table}[H]
\caption{Pool-average accuracy changes from 9 to 17 votes in the four
SuperGPQA pools, each with 3,600 questions (pp). Separate, Shared, and Exact
change follow Table~\ref{tab:allbudgetpools}. Width reduction is the
percentage decrease from separate to shared interval width; Sign identified
counts questions under Separate / Shared. Parentheses give the reasoning
effort.}
\label{tab:budgetsuper}
\centering\footnotesize
\begin{tabular}{@{}lrrrrrr@{}}
\toprule
Model & Questions & Separate & Shared & Exact change & Width reduction & Sign identified \\
\midrule
Qwen3.6-35B-A3B & 3,600 & $[-7.73,9.79]$ & $[-1.00,2.87]$ & $0.25$ & 78.0\% & 36 / 703 \\
gpt-oss-20b (high) & 3,600 & $[-9.93,12.49]$ & $[-1.43,3.62]$ & $0.19$ & 77.5\% & 88 / 833 \\
gpt-oss-20b (medium) & 3,600 & $[-12.27,15.33]$ & $[-1.76,4.38]$ & $0.29$ & 77.8\% & 94 / 1,011 \\
gpt-oss-20b (low) & 3,600 & $[-14.28,17.85]$ & $[-2.08,5.08]$ & $0.56$ & 77.7\% & 51 / 1,066 \\
\bottomrule

\end{tabular}
\end{table}

\begin{table}[H]
\caption{Budget comparisons for 3,126 model--question records in the 14
GPQA and Adaptive-Consistency pools with more than two options and 14,400
SuperGPQA records. Width reduction is the median percentage decrease in
pool-average interval width from Separate to Shared. Sign identified reports
record counts (and percentages) whose accuracy-change sign is determined by
each interval. Budgets indicate the change from fewer to more votes.}
\label{tab:budgetsummary}
\centering\footnotesize
\setlength{\tabcolsep}{2.5pt}
\begin{tabular}{@{}llrrr@{}}
\toprule
 & & & \multicolumn{2}{c}{Sign identified} \\
\cmidrule(lr){4-5}
Dataset group & Budgets & Width reduction & Separate & Shared \\
\midrule
GPQA and & $3\!\to\!9$ & 40.9\% & 1,951 (62.4\%) & 2,188 (70.0\%) \\
Adaptive-Consistency & $9\!\to\!17$ & \textbf{72.3\%} & 959 (30.7\%) & 1,717 (54.9\%) \\
 & $17\!\to\!33$ & 62.0\% & 829 (26.5\%) & 1,209 (38.7\%) \\
\midrule
SuperGPQA & $3\!\to\!9$ & 46.4\% & 3,770 (26.2\%) & 5,957 (41.4\%) \\
 & $9\!\to\!17$ & \textbf{77.7\%} & 269 (1.9\%) & 3,613 (25.1\%) \\
 & $17\!\to\!33$ & 77.4\% & 44 (0.3\%) & 1,699 (11.8\%) \\
\bottomrule
\end{tabular}
\end{table}

\end{document}